\def\paperID{0000}
\def\confName{CVPR}
\def\confYear{2024}

\newif\ifreview 
\newif\ifarxiv \newcommand{\arxiv}{\arxivtrue}
\newif\ifcamera 
\newif\ifrebuttal 

\arxiv 

\pdfoutput=1
\documentclass[10pt,twocolumn,letterpaper]{article}
\ifreview \usepackage[review]{cvpr} \fi
\ifarxiv \usepackage[pagenumbers]{cvpr} \fi
\ifrebuttal \usepackage[rebuttal]{cvpr} \fi
\ifcamera \usepackage{cvpr} \fi

\usepackage{graphicx}	
\usepackage{amsmath}	
\usepackage{amssymb}	
\usepackage{booktabs}
\usepackage{times}
\usepackage{microtype}
\usepackage{epsfig}
\usepackage[table,xcdraw,dvipsnames]{xcolor}
\usepackage{caption}
\usepackage{float}
\usepackage{placeins}
\usepackage{color, colortbl}
\usepackage{stfloats}
\usepackage{enumitem}
\usepackage{tabularx}
\usepackage{xstring}
\usepackage{multirow}
\usepackage{xspace}
\usepackage{url}
\usepackage{subcaption}
\usepackage{xcolor}
\usepackage[hang,flushmargin]{footmisc}

\ifcamera \usepackage[accsupp]{axessibility} \fi

% ...

\ifarxiv  \fi

\newcommand{\R}[1]{{%
    \textbf{%
        \ifstrequal{#1}{1}{\textcolor{red}{R#1}}{%
        \ifstrequal{#1}{2}{\textcolor{blue}{R#1}}{%
        \ifstrequal{#1}{3}{\textcolor{magenta}{R#1}}{%
        \ifstrequal{#1}{4}{\textcolor{teal}{R#1}}{%
                           \textcolor{cyan}{R#1}%
        }}}}%
    }%
}}

\usepackage{xr-hyper}

\makeatletter
\newcommand*{\addFileDependency}[1]{
  \typeout{(#1)}
  \@addtofilelist{#1}
  \IfFileExists{#1}{}{\typeout{No file #1.}}
}

\makeatother
\newcommand*{\myexternaldocument}[1]{
    \externaldocument{#1}
    \addFileDependency{#1.tex}
    \addFileDependency{#1.aux}
}

\definecolor{cvprblue}{rgb}{0.21,0.49,0.74}
\usepackage[pagebackref,breaklinks,colorlinks,citecolor=cvprblue]{hyperref}
\usepackage[capitalize]{cleveref}
\crefname{section}{Sec.}{Secs.}
\crefname{table}{Table}{Tables}
\crefname{figure}{Fig.}{Figs.}

\ifarxiv \crefname{appendix}{App.}{Apps.}
\else \crefname{appendix}{Suppl.}{Suppls.} \fi

\unless\ifarxiv \myexternaldocument{_supplementary} \fi

\usepackage{wrapfig}

\begin{document}
\title{Generating Diverse Agricultural Data for Vision-Based Farming Applications}

\author{\phantom{---}Mikolaj Cieslak\\
\phantom{---}GreenMatterAI\\
\phantom{---}{\tt\small mikolaj.cieslak@greenmatter.ai}
\and
\phantom{---}Umabharathi Govindarajan\\
\phantom{---}Blue River Technology\\
\phantom{---}{\tt\small uma.govindarajan@bluerivertech.com}
\and
\phantom{---}Alejandro Garcia\\
\phantom{---}GreenMatterAI\\
\phantom{---}{\tt\small alejandro.garcia@greenmatter.ai}
\and
\phantom{---}Anuradha Chandrashekar\\
\phantom{---}Blue River Technology\\
\phantom{---}{\tt\small anu.chandrashekar@bluerivertech.com}
\and
\phantom{------}Torsten Hädrich\\
\phantom{------}GreenMatterAI\\
\phantom{------}{\tt\small torsten.haedrich@greenmatter.ai}
\and
Aleksander Mendoza-Drosik\\
GreenMatterAI\\
{\tt\small aleksander.mendoza-drosik@greenmatter.ai}
\and
\hspace{-1cm}Dominik L. Michels\\
\hspace{-1cm}GreenMatterAI / KAUST / TU Darmstadt\\
\hspace{-1cm}{\tt\small dominik.michels@greenmatter.ai}
\and
\phantom{---}Sören Pirk\\
\phantom{---}GreenMatterAI / CAU\\
\phantom{---}{\tt\small soeren.pirk@greenmatter.ai}
\and
\hspace{0.95cm}Chia-Chun Fu\\
\hspace{0.95cm}Blue River Technology\\
\hspace{0.95cm}{\tt\small jj.fu@bluerivertech.com}
\and
\phantom{--------}Wojciech Pałubicki\\
\phantom{--------}GreenMatterAI / AMU\\
\phantom{--------}{\tt\small wojciech.palubicki@greenmatter.ai}
}

\maketitle

\begin{abstract}
\vspace{-0.35cm}
We present a specialized procedural model for generating synthetic agricultural scenes, focusing on soybean crops, along with various weeds. This model is capable of simulating distinct growth stages of these plants, diverse soil conditions, and randomized field arrangements under varying lighting conditions. The integration of real-world textures and environmental factors into the procedural generation process enhances the photorealism and applicability of the synthetic data. Our dataset includes 12,000 images with semantic labels, offering a comprehensive resource for computer vision tasks in precision agriculture, such as semantic segmentation for autonomous weed control. We validate our model's effectiveness by comparing the synthetic data against real agricultural images, demonstrating its potential to significantly augment training data for machine learning models in agriculture. This approach not only provides a cost-effective solution for generating high-quality, diverse data but also addresses specific needs in agricultural vision tasks that are not fully covered by general-purpose models.
\end{abstract}
\section{Introduction}
\label{sec:intro}
Advancements in computer vision, particularly within the agricultural domain, are increasingly reliant on the availability of diverse and accurately labeled datasets. Synthetic data emerges as a valuable resource in this context, offering several advantages:
\begin{itemize}
    \item \textit{Cost-Effectiveness and Accuracy:} Synthetic data provides labeled information that is often cheaper to acquire and less error-prone compared to manually labeled data \cite{wood2021fake, xiang2020sapien,klein2023synthetic}.
    \item \textit{Diversity and Edge Cases:} It enables the generation of diverse data, including rare edge cases, which are crucial for robust model training \cite{wrenninge2018synscapes}.
    \item \textit{Tailored Conditions:} For agriculture-specific applications, there is a need for datasets that reflect a variety of crop types at different growth stages, diseases, soil types, field arrangements, and environmental conditions as has been shown for other outdoor and indoor applications \cite{wang2021tartanvo, zhang2017physically}.
\end{itemize}

While general-purpose procedural models like Infinigen \cite{raistrick2023infinite} offer broad solutions, agricultural vision tasks demand a more specific approach. Essential factors in this domain include diverse crop types at various growth stages, disease markers, soil conditions, and specific field arrangements. Our work aims to address these needs by generating synthetic images of crops using state-of-the-art procedural models of plants and their environments in a controllable and efficient manner.

Specifically in this work, we introduce a procedural modelling approach to build virtual soybean fields which we used to generate a realistic synthetic dataset. This dataset comprises 12,000 synthetic images and encompasses:
\begin{enumerate}
    \item Multiple growth stages of soybean plants and common weeds.
    \item Variations in soil types and conditions, reflecting real-world agricultural diversity.
    \item Randomized field arrangements and weed placement, offering realistic and challenging scenarios for computer vision models.
    \item Diverse camera angles and lighting conditions to simulate different times of the day and observational perspectives.
    \item Semantic labels for crops and weeds, with the potential for other types of labels.
\end{enumerate}

To validate our approach, we compare our synthetic dataset against a corresponding set of real soybean plant images provided by Blue River Technology (\url{https://bluerivertechnology.com/}). This evaluation focuses on a semantic labeling task aimed at distinguishing between crop and weed plants, a critical step for training autonomous agricultural machinery. Our analysis involves testing models trained on various combinations of real and synthetic datasets, supplemented by cosine similarity tests and embedding visualizations to assess the similarity between synthetic and real imagery. Through this work, we demonstrate the effectiveness of our synthetic data in closely mirroring real-world conditions, thereby supporting the development of more accurate and robust computer vision models for agricultural applications.

\section{Related Work}
\label{sec:related}

The utilization of synthetic data, especially from computer graphics, has been increasingly popular in computer vision, spanning a variety of tasks and domains \cite{wood2021fake, wrenninge2018synscapes}. For an extensive overview, readers are referred to Nikolenko \cite{nikolenko2021synthetic}, who provides a comprehensive survey of this field. Below, we categorize existing work in relation to agricultural applications, generation methods, and the specific challenges they address.

The agricultural domain has seen a growing interest in synthetic data applications, but the focus has primarily been on indoor farming scenes \cite{Anagnostopoulou_2023_CVPR} or simplified outdoor scenarios \cite{tremblay2018falling, vonMarcard2018recovering}. These datasets have facilitated progress in tasks such as plant health monitoring, yield estimation, and weed detection. However, they often lack the complexity of real-world agricultural environments, such as varying crop stages, soil conditions, and field arrangements, which are critical for broad production deployments of precision agriculture.
The majority of synthetic datasets in agriculture are created using static libraries of 3D assets, which may lead to a lack of diversity and potential overfitting issues. Our approach, in contrast, leverages procedural generation, not just for object placement but also for dynamically simulating growth stages and environment. This method provides a more realistic and diverse range of scenarios, essential for training robust computer vision models for agriculture.

While datasets such as Synscapes \cite{wrenninge2018synscapes} and Infinigen \cite{raistrick2023infinite} have made significant strides in generating synthetic natural scenes, they often focus on broader landscapes or general natural objects. Our work specifically addresses the gap in generating complex agricultural scenes that include detailed plant models and varied soil types, reflecting the unique challenges faced in agricultural computer vision tasks.

In summary, while there have been significant advancements in synthetic data generation for computer vision, the application in agriculture, particularly in complex and varied outdoor environments, remains under-explored. Our work contributes to this field by providing a highly specialized and procedural approach to generating agricultural scenes, offering new opportunities for the development of advanced computer vision systems in precision agriculture.



\section{Method}
\label{sec:method}

Our method to generate synthetic images of crops is rooted in procedural modelling:
we use algorithms to create 3D models of plants, soils and other natural materials from sets of rules with configurable parameters.
Because every asset in the synthetic image is generated, we can produce infinite variation and composition, offering broad coverage of agriculturally-relevant scenes. 
The main components of our workflow are:
\begin{enumerate}
  \item plant and soil modeling,
  \item field composition and randomization,
  \item image rendering, and
  \item domain adaptation.
\end{enumerate}

The modeling, composition and randomization steps of our workflow are performed on the basis of procedural generation primitives from the open-source 3D computer graphics software Blender \cite{Blender2018}. For some plant species, requiring more specialized procedural models, we use the L-system-based L+C modeling language, implemented in the Virtual Laboratory plant modeling environment \cite{Prusinkiewicz2018L-systems}. The 3D models from this software are imported into Blender before the field composition and randomization takes place. 

As an optional step, we apply a domain-adaptation algorithm to the synthetic images. However, this depends on the availability of real data for training an image-to-image translation network, which learns to translate images from a synthetic domain to a more realistic one. 

\subsection{Plant models}
We created parameterized procedural models of soybean (\textit{Glycine max}), a grassy weed, and a broadleaf weed, which we used to generate a large set of diverse individual plants. The models are based on existing literature describing the plant's growth and development, and incorporate expert knowledge on plant morphology---provided by agronomists from Blue River Technology. Each model specifies the shoot architecture of the plant and the size and density of its organs during vegetative growth, before the appearance of reproductive organs. 

\begin{figure}[tp]
    \centering
    \includegraphics[width=\linewidth]{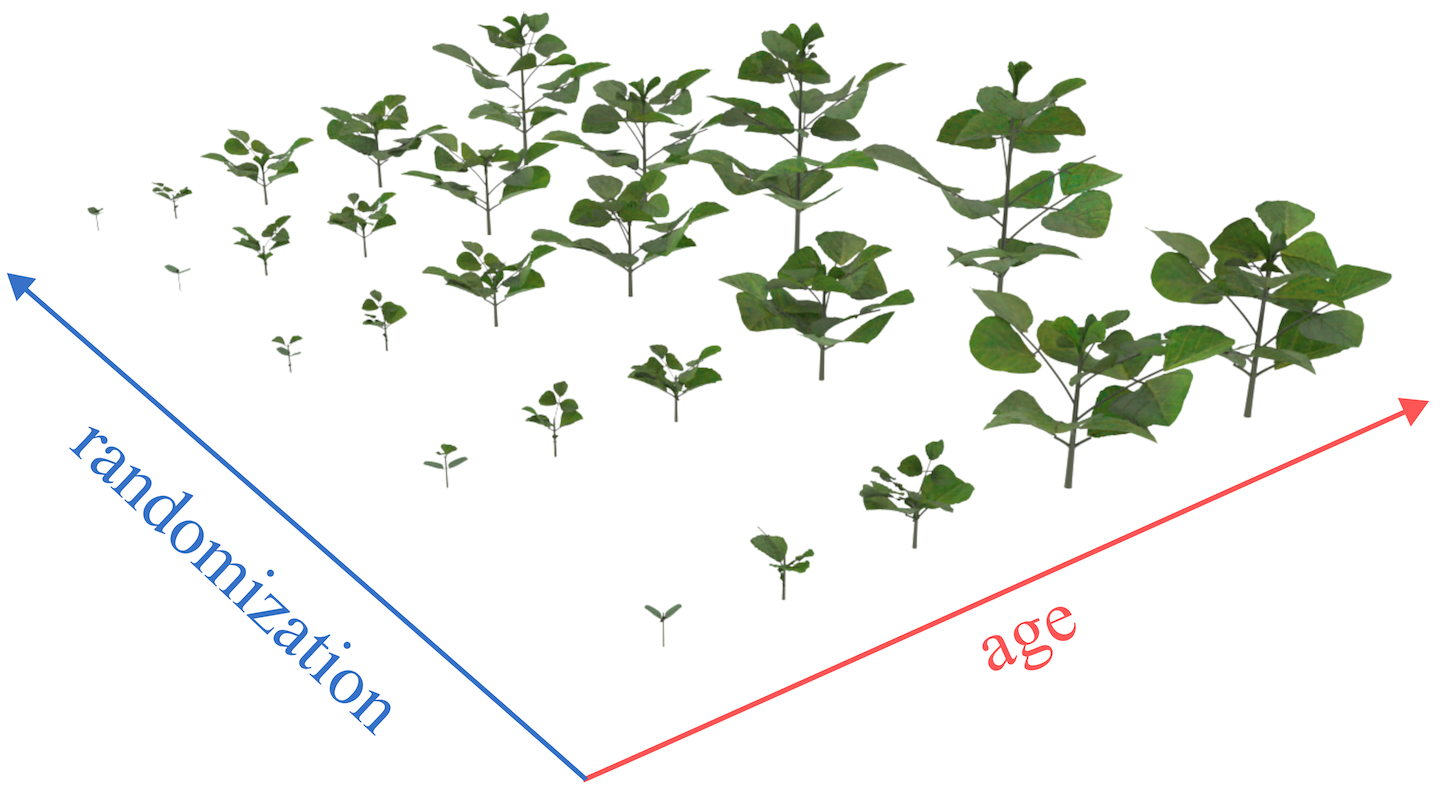}
    \caption{An example collection of 3D virtual plants generated by the procedural soybean model. Although the model has several parameters to control the plant's morphology, in practice to create fields of plants, we vary the age of the plant (x axis) and a randomization seed (y axis).}
    \label{fig:soy_growth}
    \vspace{-3mm}
\end{figure}

\textbf{Soybean model}.
Soybean is a herbaceous annual crop with growth cessation dependent on the cultivar \cite{Bernard1972soybean}. In this work, we modeled the vegetative stage only before flowering, where the plant grows from less than 15 cm to 36 cm high and produces between one and six fully opened leaves \cite{Nleya2013SoybeanGrowthStages}. Our model simulates the emergence of the stem and cotyledons, followed by the the first two unifoliate leaves, and then subsequent trifoliate leaves (see Fig.~\ref{fig:soy_growth}). Because we modeled these early developmental stages, which have a simple architecture, our soybean model is constructed using procedural generation with Blender's geometry nodes paradigm. The main stem of a single soy plant is modeled as a Bezier curve along which cotyledons, unifoliate leaves, and trifoliate leaves may be arranged. Additionally, the position of these organs along the curve determines some of their properties such as size and orientation.


\textbf{Grassy weed model}. Our model of a grassy weed is a modification of a descriptive model of the vegetative development of corn presented by Cieslak et al. \cite{Cieslak2022corn_canola}. It uses a simple rule for the distichous arrangement of the plant's leaves, which are produced on alternating sides of the stem. A different rule was used to model the production of new leaves by the main apex, simulating the emergence of successive leaves after a pre-defined period. The size and orientation of each leaf was modeled as a function of its age and position on the main stem. We defined these functions using a graphical editor, and calibrated parameters such that the model output resembled images of the real plant \cite{Cieslak2022corn_canola}. Lastly, we defined a rule to generate leaf geometry based on its simulated size and inclination angle. To model the leaf's sheath and blade, we blended between a closed and open generalized cylinder, bending and twisting its shape along the midrib. 

\textbf{Broadleaf weed model}. This model was developed using L-systems in the Virtual Laboratory \cite{Prusinkiewicz2018L-systems} on the basis of descriptions given in weed identification guides \cite{Chomas2001weedId,Fishel2020weedId}. Because we modeled higher order branching, it uses slightly more complex production rules than the grass weed model. In particular, the main apex produces lateral branches in the axil of new leaves. The rate of extension of the branches is controlled by a graphically-defined function simulating branch vigour---a phenomenological parameter modelling endogenous factors that control species-dependent branching patterns \cite{Cieslak2022corn_canola}. The geometry of the leaves, internodes, and branches is determined by their age, maximum size and position w.r.t. the parent apex. The internodes and branches are modelled as cylinders, and the leaves are modelled as open generalized cylinders, where the cross-section is defined as an open curve using a graphical editor.

\subsection{Textures and materials}
\begin{figure}[tp]
    \centering
    \includegraphics[width=\linewidth]{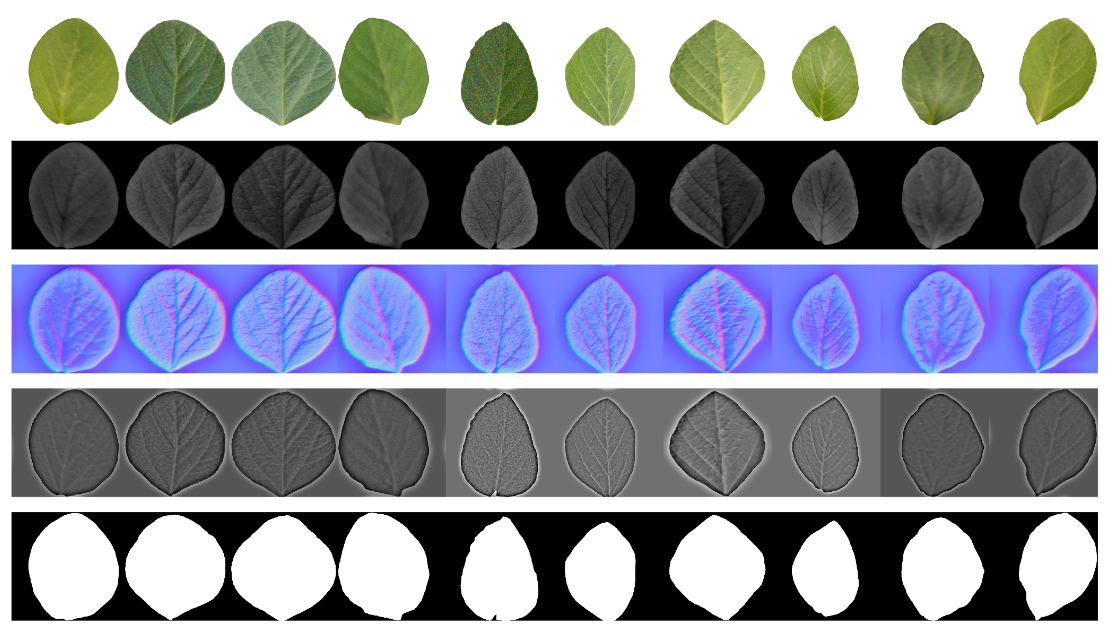}
    \caption{Texture atlases used for the soybean plants. From top to bottom, the texture atlases represent the following: diffuse/albedo map, height map, normal map, roughness map, and alpha mask map. The diffuse maps in the top row were obtained from real images through automatic segmentation, whereas the remainder were generated procedurally from their respective diffuse map .}
    \label{fig:soy_texture_atlas}
    \vspace{-3mm}
\end{figure}

To achieve photo-realistic renderings of our plant models, we developed a material pipeline that semi-automatically extracts and generates leaf textures from real images. We used the Segment Anything \cite{kirillov2023segment} model to extract diffuse maps of plant leaves that had optimal quality and orthogonal orientation towards the camera. We used these maps to generate normal, roughness, height, and alpha masks with specialized software \cite{Materialize2018}. These textures were subsequently organized into atlases for each plant. Figure \ref{fig:soy_texture_atlas} shows a texture atlas for the soybean model. Each individual leaf in a plant was assigned a random texture in such an atlas and its material properties, such as color and brightness, were then further randomized.

\subsection{Soil model and field arrangement}
The soil in our synthetic fields is modelled as a texture-mapped plane, with photorealistic material properties. We use high-quality soil textures obtained via photogammetry, as well as their displacement and roughness maps. The soil textures range in appearance from dry, cracked ground to wet, muddy soil. The displacement map is used to add depth information to the otherwise flat ground, and also to vertically displace crops, weeds and debris on the ground. Some features of our model are (1) randomization of the soil's appearance by mixing and tiling patches of its textures without any repetitions, (2) simulating variation in moisture and color temperature, and (3) addition of imprints due to farm machinery, like tire tracks.

For each image in the dataset, we specified how crops and weeds are arranged in the virtual field. To reflect cultivation practices, we set the distance between plants to be 5-10 cm and between rows of plants to be 38-76 cm \cite{Seiter2004soybean_spacing}. We also simulated seed dormancy where 10-15\% of the soybean plants did not germinate. The weeds were randomly distributed with two arrangements: between the crop rows and/or within the crop rows. The number of weeds ranged from 1 to 10 per field. Lastly, debris was distributed with a wide range of densities in a pattern controlled by fractal Perlin noise, creating clusters of material in the virtual field, but was dependent on the row spacing.




\subsection{Image rendering}

\begin{figure*}[tp]
    \centering
    \includegraphics[width=\linewidth]{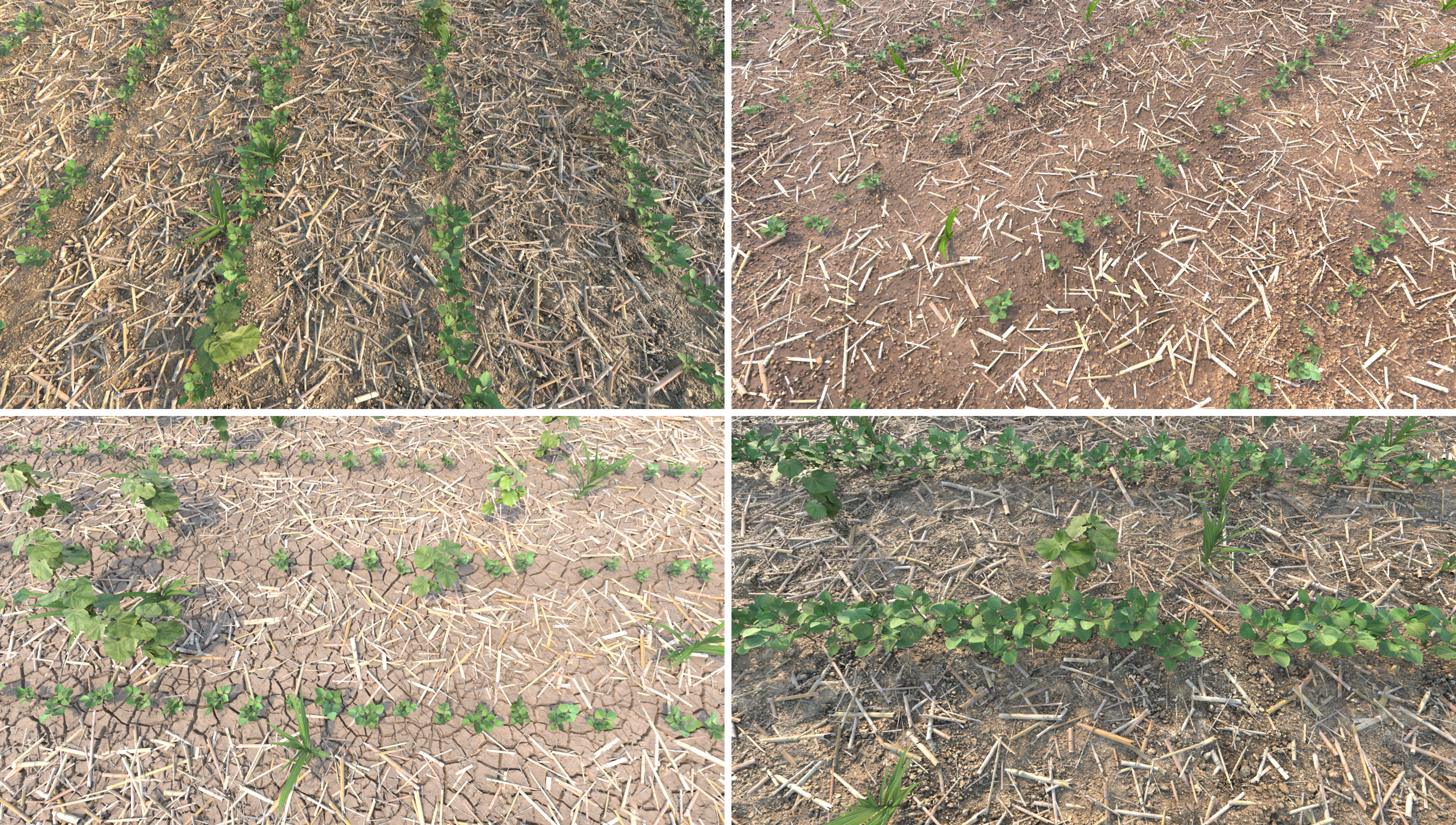}
    \caption{Selected images from our synthetic dataset, showing variation in crop growth stages, crop spacing, weed distribution, soil type, crop orientation with respect to the camera, and amount of debris.}
    \label{fig:example_synthetic_images}
    \vspace{-3mm}
\end{figure*}

We rendered images using Blender's physically-based path tracer, Cycles, which simulates the interaction between light and objects in the virtual scene. When combined with physically accurate models of light sources and optical cameras, path tracing can produce images that are hard to distinguish from photographs. Figure \ref{fig:example_synthetic_images} shows four images from our synthetic dataset. To generate accurate annotations, we disable some of Cycle's features, such as volumetric effects, motion blur, and translucency, and compute masks, surface normals and depth maps directly from the first rendering pass.

Similarly to Infinigen \cite{raistrick2023infinite}, we use Blender’s improved version of the Nishita sky model \cite{Nishita1993sky} to render the sky. The model provides an accurate simulation of atmospheric phenomena, including Rayleigh and Mie scattering, and allows us to control the position of the sun by geographical location and time of day.

\subsection{Domain adaptation}

Synthetic images can help address the need for large amounts of varied training data, but it can be challenging to generate images that close the synthetic-to-real domain gap in a way that guarantees alignment with the generated labels. To address this challenge, Fei et al. \cite{Fei2021synthetic2real} proposed using domain adaption to generate photorealistic images of crops from synthetic images with semantically consistent labels. The task was to train an image-to-image translation network on unpaired images from the synthetic domain to the real one. For this purpose they developed a semantically constrained generative adversarial network (GAN), which extends the cycle consistent adversarial network (CycleGAN) \cite{Zhu2017CycleGAN}. They used CycleGAN to generate realistic fake images but added a task-specific semantic constraint loss, which depends on the task and network model. Fei at al. \cite{Fei2021synthetic2real} showed that with this additional loss the domain-adapted images keep spatial semantics, such as plant position and size, and are aligned with the generated labels. They also showed an improvement in a fruit detection task when using the domain-adapted images instead of rendered synthetic ones.

In a similar way, we used image-to-image translation to adapt our synthetic data to the real domain.
However, we used the GAN-based Contrastive Unpaired Translation (CUT) model proposed by Park et al. \cite{Park2020CUT}. The CUT model is related to CycleGAN but replaces the cycle consistency loss with a contrastive loss on image patches. This change achieves content preservation without the additional loss function introduced by Fei et al. \cite{Fei2021synthetic2real} to CycleGAN. It does so by training a small multi-layer perceptron classifier to maximize the mutual information between a patch of the translated image and the source image. Imbusch et al. \cite{Imbusch2022synthetic2real} showed that using the CUT model is a viable approach to minimize the synthetic-to-real domain gap in robotic systems. In this work, we examined how domain-adapted images obtained with the CUT model perform in the agricultural domain.

\section{Validation}
\label{sec:validation}


\subsection{Datasets}
\begin{figure*}[tp]
    \centering
    \includegraphics[width=\linewidth]{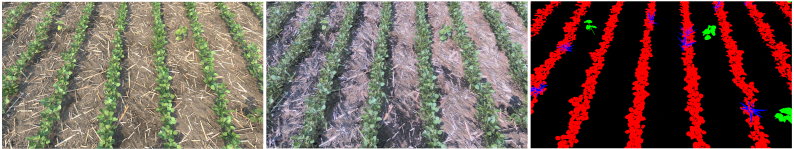}
    \caption{An example image from our synthetic dataset: (left) rendered, (middle) domain adapted, (right) generated labels, where red is crop, green is broadleaf weed, and blue is grassy weed.}
    \label{fig:example_render}
    \vspace{-3mm}
\end{figure*}

To evaluate our synthetic dataset, we divided it into two categories: (1) images rendered directly by the path tracing algorithm in Blender, and (2) images that were adapted to the real domain by passing the rendered images through an image-to-image translation network. We trained the domain-adaptation on 1,000 rendered and 1,000 real images over 400 epochs using random cropping (of size 512x512) as data augmentation. After training, we applied the translation network to all 12,000 rendered images, giving us a second synthetic dataset of 12,000 images in the synthetic-to-real domain. Figure \ref{fig:example_render} shows an example rendered image with its corresponding domain-adapted image and generated label.

\subsection{Image analysis}


A subset of 1,000 out of 12,000 images was randomly selected from each dataset. The images were passed through a pre-trained, headless ResNet-50 network, with the default ImageNet-1k weights provided by PyTorch. In addition, the images were pre-processed using the transformations associated with the pre-trained network. The resulting 2048-dimensional vectors were used to compare images with a cosine similarity test and a t-distributed Stochastic Neighbor Embedding (t-SNE) visualization.

\begin{figure}[tp]
    \centering
    \includegraphics[width=\linewidth]{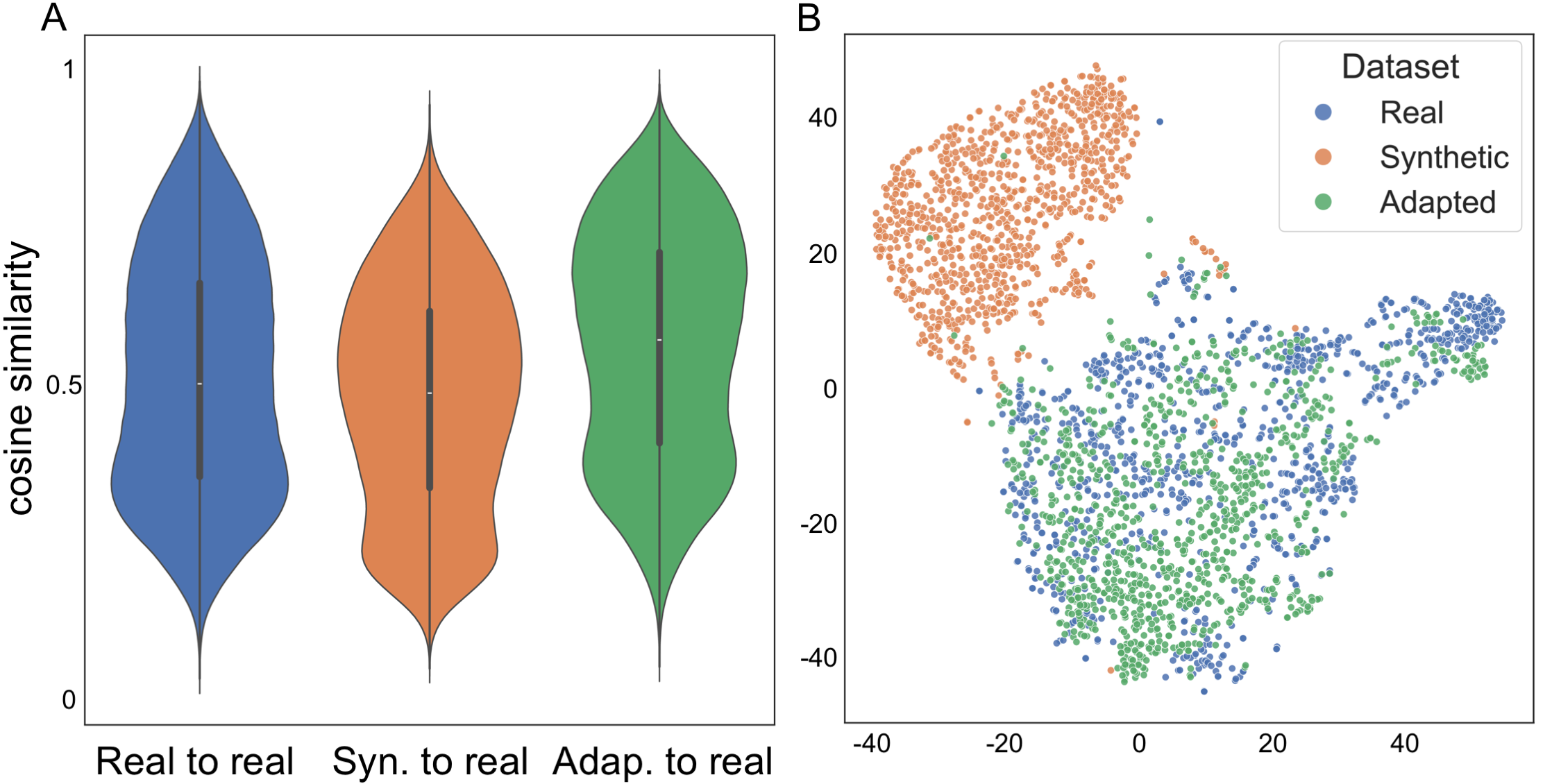}
    \caption{(A) Cosine similarity test between 1,000 images in the real and synthetic datasets. The distributions are shown between real and real images (excluding to the same image), synthetic and real images, and domain-adapted and real images. (B) t-SNE plot for the real (blue), synthetic (orange), and domain-adapted (green) datasets. Each point represents an image, which was reduced to a 2-dimensional projection from the 2048-dimensional feature vector extracted from a ResNet-50 network.}
    \label{fig:cosine_similarity_t-SNE}
    \vspace{-3mm}
\end{figure}

Figure \ref{fig:cosine_similarity_t-SNE}A shows the results of our cosine similarity test, where we computed the cosine angle (the normalized dot product) between two feature vectors for all images in the real and synthetic datasets. Violin plots are used to visualize the distributions for real images compared to all other real images, rendered images compared to real images, and domain-adapted images compared to real images. The main observations are that: (1) the median cosine similarity for the domain-adapted images is higher than for the rendered images, and (2) there is a larger proportion of domain-adapted images in the third quartile (and above) than in the rendered images. Overall, this suggests that there are many domain-adapted images that are similar to the real images. The peaks in the lower first quartile appear in all three distributions, which implies a significant number of images have features that are different from the majority of images---in terms of cosine similarity---in all three datasets.

The last part of our analysis compared the images using an embedding visualization. Figure \ref{fig:cosine_similarity_t-SNE}B shows the t-SNE plot for real and synthetic datasets. Each point in the plot corresponds to one image: blue points are real images, orange are rendered synthetic images, and green are domain-adapted synthetic images. Apart from a few outliers, we observed that the real and domain-adapted images are clustered together in the embedding. The rendered images, on the other hand, are clustered together but separated from the real images. This result further suggests that the final feature vectors of the real and domain-adapted images extracted from the pre-trained, headless ResNet-50 network are strongly similar, and the feature vectors for the rendered images are less similar. Next, we evaluate the performance of a neural network trained on our synthetic data.


\subsection{Evaluation of a crop-weed detection task}

We evaluated our synthetic dataset by training two network models to identify crops and weeds using semantic segmentation. We selected semantic segmentation as for most crop-weed detection applications, there is no need to differentiate individual plants, and object detection methods result in dense and colliding bounding boxes when the vegetation coverage is large. Wang et al. \cite{Wang2023weedmapping} tested two fully-convolutional neural networks (FCN)s, UNet and DeepLabv3+, on a similar crop-weed detection task of real images taken from an unoccupied aerial vehicle (UAV), and found the models performed well in terms of the intersection-over-union (IoU) metric. We chose to evaluate our synthetic dataset in the same way using DeepLabv3 implemented in PyTorch's torchvision library \cite{torchvision2016} and Segformer \cite{xie2021segformer} implemented in HuggingFace's Transformers library \cite{wolf-etal-2020-transformers}.

We performed three experiments: (1) we varied the ratio of synthetic to real images in the training data and tested the models on a hold-out dataset of 3,891 real images of soybean fields at various growth stages, (2) we tested the same models from (1) on a dataset of cotton fields (without training for the new crop), and (3) we set the number of real images to 5,000 and varied the amount of synthetic data, testing on the hold-out soybean dataset. For all experiments, we used the ResNet-101 backbone for DeepLabv3 and the MIT-B0 backbone for Segformer. Both models were pre-trained with ImageNet-1k weights, and we applied four data augmentation techniques: translate and reflect, scale and reflect, flip left-right, and contrast adjustment.

\begin{table*}[t]
\centering
\caption{Mean IoU values for weed and crop classes with different ratios of synthetic images to real ones. All results were tested on a hold-out dataset of 3,981 real images of soybean fields. Abbreviations: SegF = Segformer, DeepL = DeepLabv3, Syn = rendered synthetic images, Adap = domain-adapted synthetic images.}
\label{tab:experiment1}
\begin{tabular}{l|cccc|ccccc}
\toprule
\textbf{Training dataset} & \multicolumn{4}{c|}{\textbf{mean IoU - Weed}} & \multicolumn{4}{c}{\textbf{mean IoU - Crop}} \\
\textbf{syn : real ratio} & \textbf{SegF} & \textbf{SegF} & \textbf{DeepL} & \textbf{DeepL} & \textbf{SegF} & \textbf{SegF} & \textbf{DeepL} & \textbf{DeepL} \\
 & \textbf{Syn} & \textbf{Adap} & \textbf{Syn} & \textbf{Adap} & \textbf{Syn.} & \textbf{Adap} & \textbf{Syn} & \textbf{Adap} \\
\midrule
12k : 0 &  0.0586 & 0.0652 & 0.0507 & 0.0574 & 0.5970 & 0.4461 & 0.3664 & 0.3858  \\
10k : 2k (5:1) &  0.3058 & 0.2808 & 0.3084 & 0.2807 & 0.8011 & 0.7928 & 0.7598 & 0.7513 \\
8k : 4k (2:1) &  0.3417 & 0.3047 & 0.3223 & 0.3017 & 0.8078 & 0.7993 & 0.7646 & 0.7568 \\
6k : 6k (1:1) &  0.3415 & 0.3170 & 0.3341 & 0.3245 & 0.8099 & 0.8022 & 0.7671 & 0.7629 \\
4k : 8k (1:2) &  0.3527 & 0.3250 & 0.3386 & 0.3245 & 0.8119 & 0.8042 & 0.7674 & 0.7629 \\
2k : 10k (1:5) &  0.3559 & 0.3385 & 0.3421 & 0.3323 & 0.8127 & 0.8071 & 0.7680 & 0.7658 \\
12k : 12k  (1:1) &  0.3673 & 0.3418 & 0.3471 & 0.3290 & 0.8160 & 0.8095 & 0.7674 & 0.7653 \\
0 : 12k &  0.3519 & 0.3519 & 0.3381 & 0.3381 & 0.8126 & 0.8126 & 0.7684 & 0.7684 \\
\bottomrule
\end{tabular}
\vspace{-2mm}
\end{table*}

\begin{figure}
  \centering
  \includegraphics[width=1.0\linewidth]{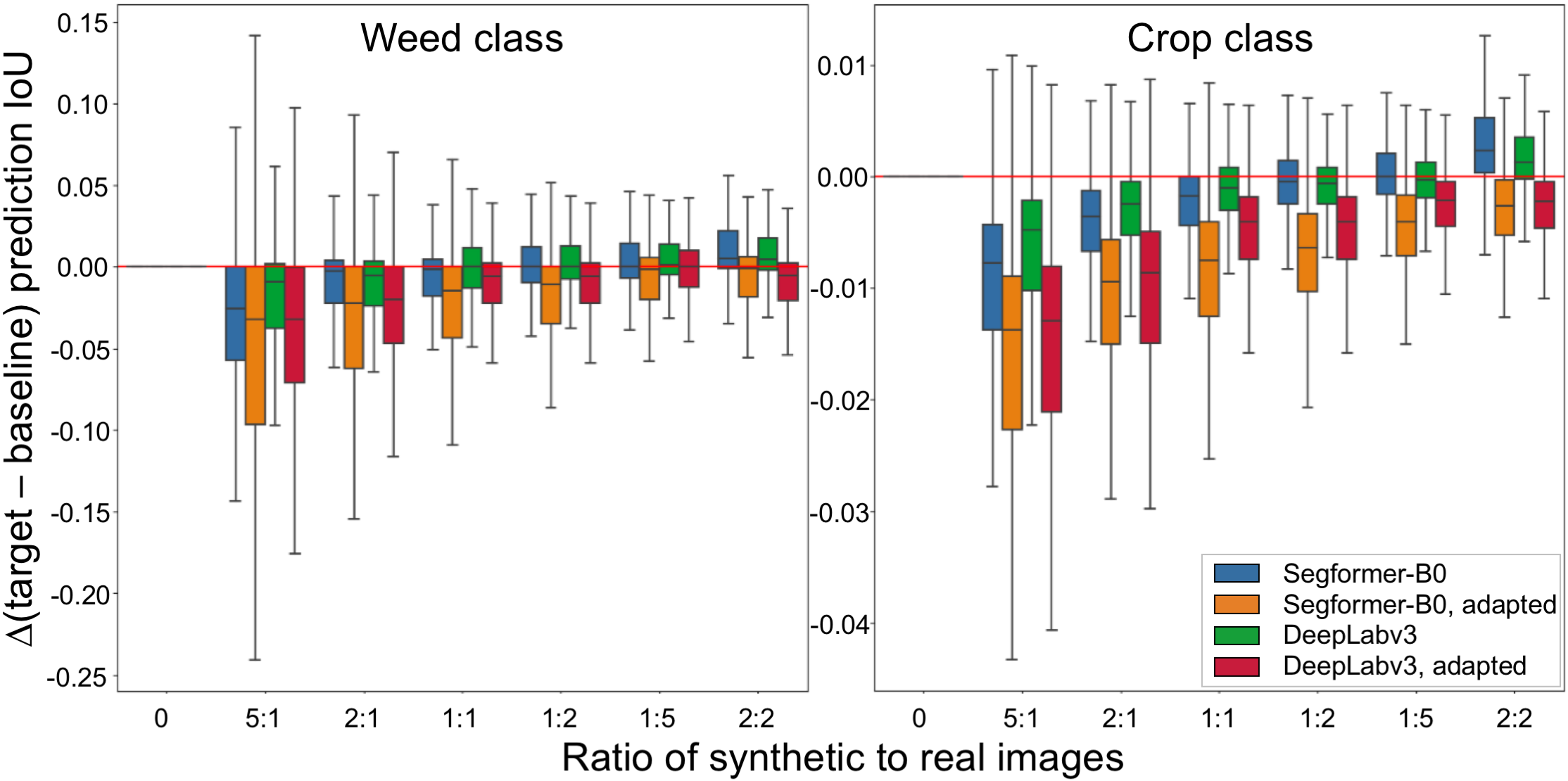}
  \caption{Box plots showing the distribution of IoU values for weed and crop classes for both models. $\Delta$ is the difference between the target IoU and the baseline IoU, which is from the model trained on 12,000 real images only.}
  \label{fig:mIoURatioExp1}
  \vspace{-3mm}
\end{figure}

Table \ref{tab:experiment1} and Fig.~\ref{fig:mIoURatioExp1} show IoU values for weed and crop classes as we varied the ratio of synthetic to real images used for training. We split the dataset to 90\% training and 10\% validation out of 12,000 images in total. The results show that training the models on synthetic images alone under performs compared to combining real and synthetic images. However, combining synthetic images with even a small number of real images provides a boost in performance. Anderson et al. \cite{Anderson2022SyntheticCars} observed a similar result for a segmentation task in a car manufacturing application. Our data shows that at a 1:2 ratio of 4,000 synthetic and 8,000 real images, both models perform as well as (or better than) a real dataset with 12,000 images. In addition, combining all synthetic and all real images, resulting in a training set size of 24,000 images, outperforms training on 12,000 real images for both classes and both models. 

\begin{figure*}
  \centering
  \includegraphics[width=\textwidth]{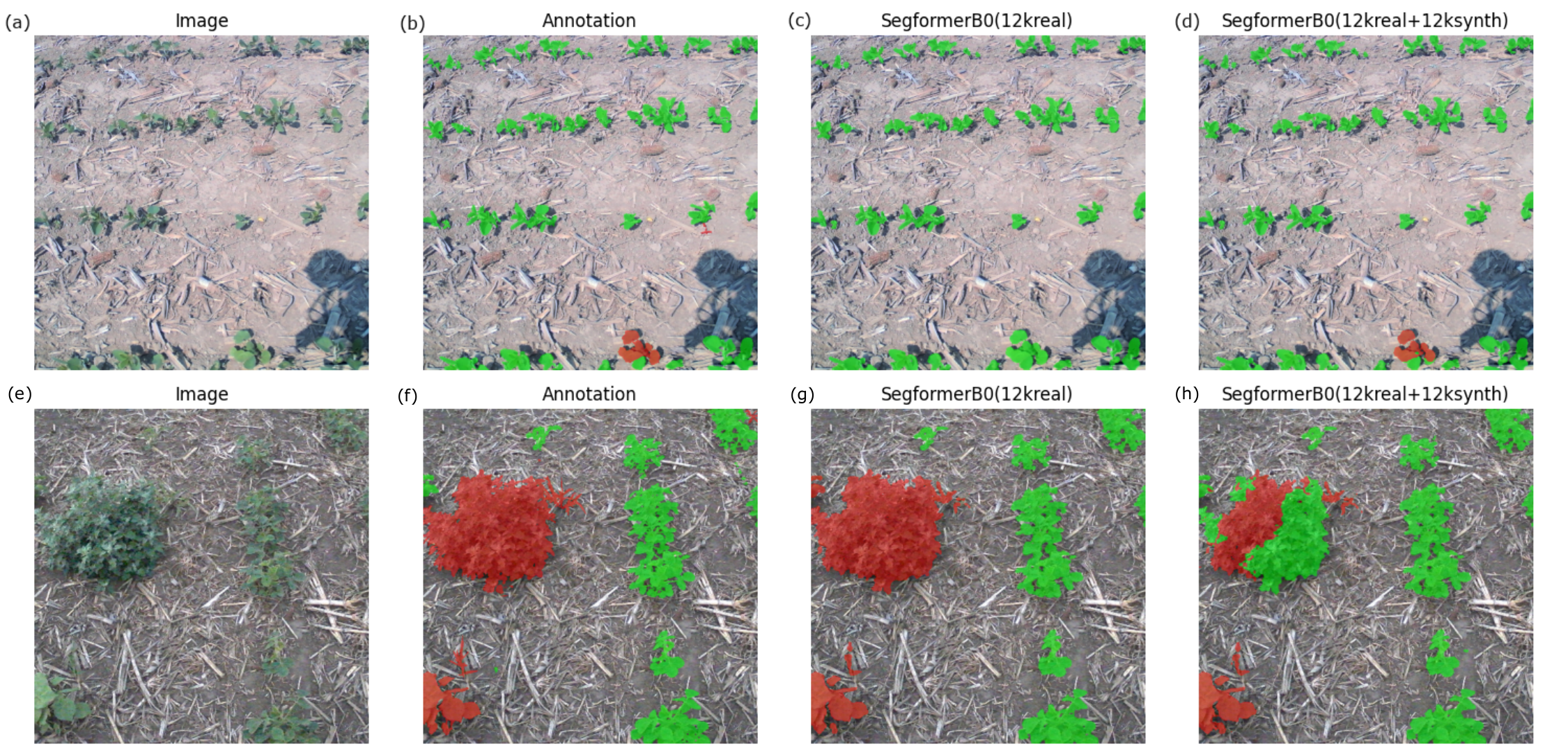}
  \caption{Two example images from the real dataset, their manually-annotated labels, and two  Segformer-based predictions: (1) trained on 12,000 real images and (2) trained on a combined 12,000 real and 12,000 synthetic images. The labels are overlaid on the images with green used for crops and red for weeds.}
  \label{fig:example_annotations}
  \vspace{-3mm}
\end{figure*}

Figure~\ref{fig:example_annotations} shows two examples of real images, their annotations and the predicted labels from the fine-tuned Segformer model. The examples were selected to highlight our observed differences in predicted labels between the model trained on 12,000 real labels and the one trained on 12,000 real and 12,000 synthetic images. The example image shown in Figure~\ref{fig:example_annotations}(a-d) shows that the model trained with synthetic images correctly predicts the weed close to the most-bottom row of crops (Fig.~\ref{fig:example_annotations}d), whereas the model trained on real images (Fig.~\ref{fig:example_annotations}c) incorrectly labels the weed as crop. Figure~\ref{fig:example_annotations}(e-h) shows a case where the model trained on the combined dataset incorrectly labels a weed partially as crop (Fig.~\ref{fig:example_annotations}h), which was correctly identified by the model trained on real data only (Fig.~\ref{fig:example_annotations}g). Please note that this particular weed species was not modeled with our synthetic generation pipeline. This might indicate that the synthetic dataset helps to improve the labelling of modeled weeds growing close to the crop, but reduces performance on weeds that were not included in the synthetic dataset.

\begin{wrapfigure}{r}{0.2\textwidth}
  \vspace{-2mm}
  \centering
  \includegraphics[width=\linewidth]{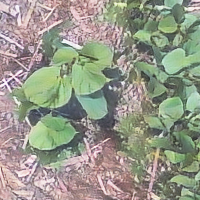}
  \caption{Close-up of domain-adapted image.}
  \vspace{-2mm}
  \label{fig:domain-adapted-green-blobs}
\end{wrapfigure}

One surprising result of the first experiment is that the domain-adapted images do not increase model performance. Since our analysis showed that domain-adapted images are more similar to the real images than the synthetic ones, we expected better mean IoU values for domain-adapted images only compared to synthetic images only. In addition, Imbusch et al. \cite{Imbusch2022synthetic2real} reported improved mean IoU values for CUT GAN translated images compared to their rendered synthetic images. There may be several possible explanations, including (1) our image analysis used a trained network with ImageNet-1k weights so the feature vector may not capture features relevant for training towards this particular vision task, and (2) the CUT GAN-processed images contain artifacts that break the semantic consistency between the images and the generated labels (after all this method does not use semantic constraint loss such as in the work of Fei et al. \cite{Fei2021synthetic2real}). For example, we observed that the CUT GAN model occasionally added green patches of pixels in regions of the image not labeled semantically as plants, which did not look like plants, and therefore could introduce inconsistencies between the images and the generated labels (see Fig.~\ref{fig:domain-adapted-green-blobs}).
These types of hallucinations are potentially caused by a mismatch between features present in the real images that are missing from our synthetic dataset. Specifically, for images containing high vegetative or soil debris coverage the CUT GAN model may have difficulty matching information between images of real plants and synthetic ones.


Table \ref{tab:experiment2} shows the results of testing the same models from our first experiment on real images of cotton fields. In other words, we tested our soybean-trained models on images of a different crop to test how resilient the models are to out-of-distribution test data. We observed a similar boost in performance for both models except at a lower ratio of 2:1, i.e., 8,000 synthetic images and 4,000 real images. This suggests that our synthetic dataset may represent features found in both the real soybean and cotton images even though it was generated specifically for soybean. It may also suggest that the combination of real and synthetic datasets generalizes better than separate real or synthetic datasets. This is indicated by the results of the different-ratio experiments shown in Table~\ref{tab:experiment1} where the mean IoU for both models improves in experiments starting from a ratio of 2:1 for the weed class and 5:1 for the crop class compared to the real training data baseline. In summary, the increased performance of models trained with higher ratios of synthetic to real training data on cotton field test data may indicate better generalization to out-of-distribution features compared to models trained with real data only.

\begin{table}[t]
\centering
\caption{Mean IoU values for weed and crop classes with different ratios of synthetic images to real ones trained on soybean crop but tested on a cotton. Abbreviations: SegF = Segformer, DeepL = DeepLabv3, Syn = rendered synthetic images.}
\label{tab:experiment2}
\begin{tabular}{l|cc|cc}
\toprule
\textbf{Train dataset} & \multicolumn{2}{c|}{\textbf{mIoU - Weed}} & \multicolumn{2}{c}{\textbf{mIoU - Crop}} \\
\textbf{syn : real ratio} & \textbf{SegF} & \textbf{DeepL} & \textbf{SegF} & \textbf{DeepL} \\
\midrule
12k : 0 &  0.0427 & 0.0359 & 0.4626 & 0.2904 \\
10k : 2k  (5:1) &  0.1203 & 0.1172 &  0.7232 & 0.6355  \\
8k : 4k (2:1) &  0.1439 & 0.1348 &  0.7256 & 0.6718  \\
6k  : 6k  (1:1) &  0.1446 & 0.1370 &  0.7214 & 0.6581 \\
4k  : 8k  (1:2) &   0.1549 & 0.1409 &  0.7285 & 0.6559 \\
2k  : 10k  (1:5) &  0.1553 & 0.1430 &  0.7155 & 0.6330 \\
12k  : 12k  (1:1) &  0.1667 & 0.1411 &  0.7268 & 0.6412 \\
0  : 12k   &  0.1429 & 0.1353 &  0.7173 & 0.6289 \\
\bottomrule
\end{tabular}
\vspace{-2mm}
\end{table}

In the last experiment, we kept the number of real images constant at 5,000, but increased the number of synthetic images in the training dataset. Table \ref{tab:experiment3} reports the mean IoU per class for both models.
We observed a slight linear increase in performance when adding more synthetic images for both classes and models. The results suggest that adding synthetic data to a training dataset is a good approach when real-annotated data is scarce. In particular, synthetic data could be useful in training a model before fine-tuning with real data. 


\begin{table*}[t]
\centering
\caption{Mean IoU values for weed and crop classes. The models were trained with a constant number of real images but augmented with different numbers of synthetic images. Testing was done on a hold-out dataset of 3,981 real images of soybean fields. Abbreviations: SegF = Segformer, DeepL = DeepLabv3, Syn = rendered synthetic images, Adap = domain-adapted synthetic images.}
\label{tab:experiment3}
\begin{tabular}{l|cccc|ccccc}
\toprule
\textbf{Training dataset} & \multicolumn{4}{c|}{\textbf{mean IoU - Weed}} & \multicolumn{4}{c}{\textbf{mean IoU - Crop}} \\
 & \textbf{SegF} & \textbf{SegF} & \textbf{DeepL} & \textbf{DeepL} & \textbf{SegF} & \textbf{SegF} & \textbf{DeepL} & \textbf{DeepL} \\
 & \textbf{Syn} & \textbf{Adap} & \textbf{Syn} & \textbf{Adap} & \textbf{Syn.} & \textbf{Adap} & \textbf{Syn} & \textbf{Adap} \\
\midrule
10k synth + 5k real & 0.3351 & 0.3139 & 0.3269 & 0.3043 & 0.8080 & 0.8017 & 0.7640 & 0.7556  \\
8k synth + 5k real & 0.3285 & 0.3119 & 0.3263 & 0.3010 & 0.8071 & 0.8022 & 0.7660 & 0.7578  \\
6k synth + 5k real & 0.3341 & 0.3144 & 0.3303 & 0.3117 & 0.8073 & 0.8012 & 0.7659 & 0.7596  \\
4k synth + 5k real & 0.3264 & 0.3035 & 0.3312 & 0.3118 & 0.8057 & 0.7984 & 0.7652 & 0.7597  \\
2k synth + 5k real & 0.3318 & 0.2971 & 0.3277 & 0.3107 & 0.8056 & 0.7975 & 0.7649 & 0.7598  \\
0 synth + 5k real & 0.3237 & 0.3237 & 0.3155 & 0.3155 & 0.8036 & 0.8036 & 0.7638 & 0.7638  \\
\bottomrule
\end{tabular}
\vspace{-2mm}
\end{table*}



\section{Conclusion}
\label{sec:conclusion}

We presented a fully automatic, procedural workflow to generate large-scale synthetic datasets in the agricultural domain. On this basis, we generated a synthetic dataset of 12,000 images of soybean fields and used it to create a second set of 12,000 domain-adapted synthetic images. We analyzed the synthetic images by performing a cosine similarity test and a t-SNE embedding visualization, showing the synthetic images are similar to the real ones. We then evaluated both datasets by training a Segformer and DeepLabv3 model on various combinations of real and synthetic datasets, and tested the models on hold-out datasets of real images of soybean and cotton fields. We found that, for our crop-weed image segmentation task, synthetic images were an effective data augmentation strategy, confirming similar results from non-agricultural domains \cite{Anderson2022SyntheticCars, Imbusch2022synthetic2real}.

In contrast to Imbusch et al. \cite{Imbusch2022synthetic2real}, however, we found that the models trained on rendered synthetic images or the domain-adapted counterparts showed similar mean IoU values for our segmentation task. A possible reason for this discrepancy may be the difference in scene complexity---natural scenes like fields of crops are more difficult to model than cluttered tabletop scenes of man-made objects---where the CUT GAN-based domain adaption method may not perform as well. In addition, Imbusch et al \cite{Imbusch2022synthetic2real} trained a network model for semantic segmentation from scratch on 450,000 images, whereas we used pre-trained models with ImageNet-1k weights and mixed up to 24,000 real and synthetic images. Fei et al. \cite{Fei2021synthetic2real} did report an improvement in a fruit detection task with domain-adapted images but with low numbers of training images---less than 50---and marginal improvement for slightly more images. 

We conclude that training vision models with splits of synthetic and real data in the agricultural domain can lead to the same or better performance compared to real training datasets of the same size. Further, we found that models trained on combinations of real and synthetic datasets of crop fields generalize better than models trained on real datasets to other images of real fields. This seems to be especially useful for training resilient models for agriculture-specific vision tasks considering the large amount of variability in crop appearance, management strategies, geographical and seasonal differences. Next, we plan on evaluating the characteristics of our synthetic data to find features that contribute to model performance and identify edge-cases where real data is lacking and synthetic data can provide a boost in performance. One way forward is to use the Deep Generative Ensemble framework proposed by van Breugel et al. \cite{vanBreugel2023DGE}, which helps to approximate the posterior distribution of generative models, identifying low-density regions of the original data.



{\small
\bibliographystyle{ieeenat_fullname}
\bibliography{11_references}
}

\ifarxiv \clearpage \appendix 

 \fi

\end{document}


\title{\paperTitle}
\author{\authorBlock}
\maketitlesupplementary

\appendix


{\small
\bibliographystyle{ieeenat_fullname}
\bibliography{11_references}
}